\pdfoutput=1

\documentclass[11pt]{article}

\usepackage[final]{acl}

\usepackage{times}
\usepackage{latexsym}

\usepackage[T1]{fontenc}

\usepackage[utf8]{inputenc}

\usepackage{microtype}

\usepackage{inconsolata}

\usepackage{graphicx}
\usepackage{algorithm}
\usepackage{algpseudocode}
\usepackage{amsmath}      
\usepackage{amssymb}
\usepackage{adjustbox}
\usepackage{booktabs}
\usepackage{multirow}
\usepackage{xcolor}
\usepackage{colortbl}
\newtheorem{theorem}{Theorem}[section]
\newtheorem{proposition}[theorem]{Proposition}
\title{Two-Stage Regularization-Based Structured Pruning for LLMs}

\author{
  \textbf{Mingkuan Feng}\textsuperscript{\rm 1}\thanks{\quad Equal contribution}, \textbf{Jinyang Wu}\textsuperscript{\rm 1}\footnotemark[1], \textbf{Siyuan Liu}\textsuperscript{\rm 2}\footnotemark[1], \textbf{Shuai Zhang}\textsuperscript{\rm 1}\thanks{\quad Corresponding Authors},\\
  \textbf{Hongjian Fang}\textsuperscript{\rm 3}\footnotemark[2],
  \textbf{Ruihan Jin}\textsuperscript{\rm 1}, \textbf{Feihu Che}\textsuperscript{\rm 3}, \textbf{Pengpeng Shao}\textsuperscript{\rm 3}, \textbf{Zhengqi Wen}\textsuperscript{\rm 3}\footnotemark[2], \textbf{Jianhua Tao}\textsuperscript{\rm 1,\rm3}\footnotemark[2]\\
  \textsuperscript{\rm 1}Tsinghua University\\
  \textsuperscript{\rm 2}Peking University\\
  \textsuperscript{\rm 3}Beijing National Research Center for Information Science and Technology\\
    \texttt{fmk24@mails.tsinghua.edu.cn}\\
    }

\begin{document}
\maketitle
\begin{abstract}
The deployment of large language models (LLMs) is largely hindered by their large number of parameters. Structural pruning has emerged as a promising solution. Prior structured pruning methods directly remove unimportant parameters based on certain metrics, which often causes knowledge loss and necessitates extensive retraining. To overcome this, we introduce a novel pruning method \textbf{TRSP}: \textbf{T}wo-Stage \textbf{R}egularization-Based \textbf{S}tructured \textbf{P}runing for LLMs. Specifically, we multiply the output of each transformer layer by an initial learnable weight and iteratively learn these weights by adding their $\ell_1$-norm as a regularization term to the loss function, serving as the first-stage regularization. Subsequently, we apply additional regularization to the difference between the output and input of layers with smaller weights, encouraging the shift of knowledge to the preserved layers. This serves as the second-stage regularization. TRSP retains more knowledge and better preserves model performance than direct parameter elimination. Through extensive experimentation we show that TRSP outperforms strong layer-wise structured pruning methods without requiring retraining. As a layer-wise pruning method, it delivers notable end-to-end acceleration, making it a promising solution for efficient LLM deployment. Code is available at \url{https://github.com/fmk345/TRSP}.
\end{abstract}

\section{Introduction}
Large language models (LLMs) have made remarkable progress in natural language processing~\cite{yang2024qwen2,wu2024beyond,guo2025deepseek,wu2025templaterl,wu2025pandora,liu2025phd,liu2025empowering,liu2025better,liu2025segmentation,jin2025radialrouter,jin2026exploring,wu2026astar}. However, their large scale makes real-world deployment challenging. There is an urgent need for techniques that can enhance the compactness and computational efficiency of LLMs while preserving their language modeling capabilities.

Structured pruning is a method used to simplify neural networks by removing unnecessary or redundant parameters~\cite{xia2024sheared,an2024fluctuation,feng2025dress}. Structured pruning is categorized into channel-wise pruning~\cite{ma2023llm,ashkboos2024slicegpt} and layer-wise pruning~\cite{song2024sleb}. Channel-wise pruning operates at the row or column level of parameter matrices. Layer-wise pruning operates at the level of entire transformer layers thereby offering a simpler approach compared to channel-wise pruning~\cite{chen2024streamlining, men2024shortgpt}.

However, existing layer-wise pruning methods have a certain limitation. They consistently first compute the importance of each transformer layer using a designed criteria, prune unimportant layers, and then fine-tune the pruned model to compensate for performance degradation caused by pruning. However, even unimportant layers can hold valuable knowledge~\cite{dettmers2022gpt3,yin2024outlier,an2025systematic}. This sequential process of selecting and then directly pruning layers does not handle the important knowledge contained in the layers that are to be pruned, resulting in its direct loss. The performance drop requires substantial retraining for recovery, leading to considerable computational overhead~\cite{ma2023llm}.

To address this, we present TRSP that first applies two-stage regularization and then prune. The first regularization process iteratively learns layer weights. The second regularization process dynamically transfers valuable knowledge from the layers to be pruned to the remaining layers in advance, greatly reducing the knowledge loss caused by pruning. 

\begin{figure*}[t]
	\centering
    \includegraphics[width=.99\textwidth]{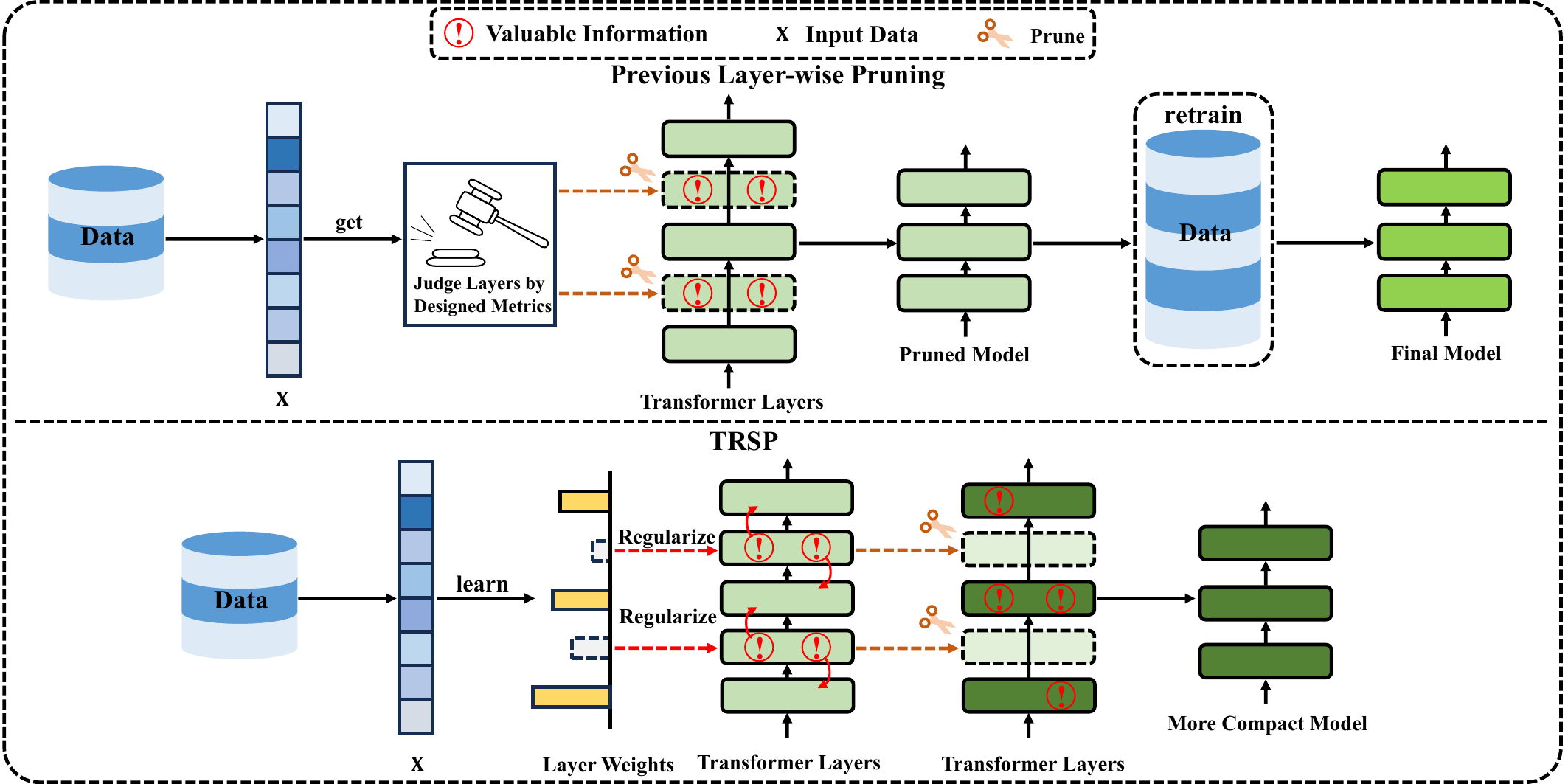}
    \caption{A comparison between existing layer-wise structured pruning methods and TRSP. Deeper blue layer represents greater performance impact, the taller cylinder represents larger data volume.}
  \label{figure1}
\end{figure*}

The comparison of TRSP with existing layer-wise pruning approaches is shown in Figure \ref{figure1}. \underline{\textit{First}}, we sample a small portion of data from standard benchmark datasets randomly. Given the limited scale of the selected data, the computational overhead incurred during the two-stage regularization is significantly reduced. \underline{\textit{Second}}, we multiply the output of each transformer layer by an initial learnable weight and iteratively learn these weights by incorporating their $\ell_1$-norm as a regularization term in the loss function. This serves as the first stage regularization. \underline{\textit{Third}}, we apply regularization ($\ell_1$-norm or $\ell_2$-norm) to the difference between the outputs and inputs of the layers with smaller weights, forcing important knowledge to be transferred to the remaining layers which significantly reduces the performance decline caused by parameter removal. Thus the model can maintain good language modeling capability. This serves as the second stage regularization. \underline{\textit{Finally}}, we prune the layers with smaller weights. Comprehensive experiments demonstrate that TRSP substantially outperforms strong layer-wise pruning methods in generation tasks and zero-shot tasks across different pruning ratios, while also significantly improving end-to-end acceleration. The main contributions of TRSP are summarized as follows:
\begin{itemize}
\item\textbf{Retention of Knowledge: }TRSP effectively mitigates knowledge loss by progressively applying two-stage regularization, followed by structured pruning. This systematic approach ensures that model performance is well preserved throughout the compression process, without the need for costly retraining.
\item\textbf{Effectiveness: }TRSP outperforms strong layer-wise pruning methods in generation and zero-shot tasks. The pruned model demonstrates a considerable acceleration.
\item\textbf{Minimal Cost: }The data required for two-stage regularization is minimal and TRSP is retraining free after pruning.
\end{itemize}

\section{Related Work}
\subsection{Model Pruning}
Model Pruning aims to improve model efficiency by sparsification or parameter removal~\cite{lecun1989optimal,hassibi1993optimal,han2015learning,liu2017learning}. Several studies employ unstructured~\cite{kurtic2022optimal,zhang2024plugandplay,xu2024besa} and structured pruning~\cite{xia2024sheared,yang2024laco,gao2024disp}.

Unstructured pruning zeros individual neurons according to their importance such as SparseGPT~\cite{frantar2023sparsegpt}, SpQR~\cite{dettmers2024spqr}, Pruner-Zero~\cite{dong2024pruner}, and Wanda~\cite{sun2024a}. Their main advantage is flexibility. However, they need dedicated hardware to accelerate~\cite{xia2023flash}, and are not able to retrain on downstream tasks.

Structured pruning methods can be categorized by granularity into channel-wise pruning~\cite{ashkboos2024slicegpt} and layer-wise pruning~\cite{kim2024shortened}. 

Channel-wise pruning methods create a metric to assess the significance of channels in the parameter matrices of LLMs, and then remove the less significant ones. 

Layer-wise pruning methods treat entire layers as the basic units for pruning~\cite{chen2025dlp}. For instance, SLEB~\cite{song2024sleb} iteratively removes entire transformer layers by evaluating their impact on the model's final loss, ShortGPT~\cite{men2024shortgpt} thinks high similarity between layers means redundancy, LaCo~\cite{yang2024laco} uses layer collapse to prune, Shortened LLaMA~\cite{kim2024shortened} prunes layers in one-shot based on their importance. In this paper, we focus on layer-wise pruning. Prior layer-wise pruning approaches suffer from the drawback that the less important layers may still carry critical knowledge, and pruning them often causes knowledge loss. Discovering a way to reshape knowledge distribution prior to pruning could potentially alleviate knowledge loss.
\subsection{Regularization}
In machine learning, regularization plays a vital role in controlling overfitting~\cite{santos2022avoiding} and identifying informative features~\cite{tibshirani1996regression}, and has been extensively studied~\cite{hoerl1970ridge,poggio1987computational,balestriero2022effects}. 

The $\ell_1$-norm tends to enforce compact representations by eliminating certain parameters, whereas the $\ell_2$-norm favors stability and continuity~\cite{boyd2004convex}. Both can alter the underlying structure and representation of the data~\cite{han2015learning,tao2023structured}. Inspired by this insight, regularization can be leveraged to migrate critical knowledge from pruned layers to preserved layers, thereby enhancing model performance.

\begin{algorithm}[H]
   \caption{TRSP algorithm.}
   \label{algorithm1}
\begin{algorithmic}[1]
   \State {\bfseries Input:} selected data: $\mathbf{X}$, number of layers: $l$, initial model $\mathbf{W}$, number of layers to prune: $n$, layer weights: $S$, set of pruned layers: $P$, norm type: $\text{flag}$.
   \State Initialize each layer weight in $S$ to 1.
   \State $P \leftarrow \emptyset$, $MinS \leftarrow 1e9$
   \For{$i=0$ {\bfseries to} $n-1$}
     \State$S\leftarrow learnWeights(\mathbf{W},S,\mathbf{X})$
    \For{$j=0$ {\bfseries to} $l-i-1$}
       \If{$S[j]<MinS$} 
       \State $MinS\leftarrow S[j]$
       \State $MinS\_id\leftarrow j$
   \EndIf
    \EndFor
    \State $\mathbf{W} \leftarrow mask(\mathbf{W},MinS\_id)$
        \State $P \leftarrow P \bigcup \{MinS\_id\}$
\EndFor
\State $\mathcal{L}_{\text{sum}}\leftarrow \mathcal{L}(\mathbf{W}, \mathbf{X})$
   \For{$i=0$ {\bfseries to} $\mathrm{sizeof}(P)$}
    \State$\mathcal{L}_{\text{sum}}\leftarrow\mathcal{L}_{\text{sum}}+regularized(\mathbf{W}[P[i]])$
   \EndFor
\State update $\mathbf{W}$ using backpropagation algorithm
\State $\mathbf{W} \leftarrow Prune(\mathbf{W},P)$
   \end{algorithmic}\end{algorithm}

\section{Methodology}
\label{headings}
The complete TRSP procedure is outlined in Algorithm \ref{algorithm1} and Figure~\ref{figure1}. The TRSP framework involves four key stages: \textbf{(1) Prepare data: }Select a small amount of data for the following two-stage regularization. \textbf{(2) Learn layer weights: }Iteratively learn the weights of each layer by incorporating their $\ell_1$-norm as a regularization term in the loss function. \textbf{(3) The second stage regularization: }Apply regularization to the difference between the output and input of layers with smaller weights, facilitating knowledge transfer. \textbf{(4) Pruning: }Removes the layers that were regularized in the previous step.

\subsection{Prepare Data}
The data is drawn from standard benchmarks, including Alpaca \cite{alpaca}, WikiText-2 \cite{merity2016pointer}, PTB \cite{marcus1993building}, and C4 \cite{raffel2020exploring}. For example, 128 instances are randomly drawn from the WikiText-2 training set for layer weight learning via regularization and the second stage regularization.
\subsection{Learn Layer Weights}
\label{3.2}
This is the first stage regularization. To better understand our paper, we first define some notations. $\mathbf{W}$ represents the initial model. Let \( p \) be the pruning ratio, indicating that \( p\% \) of model layers will be pruned. The number of layers in the initial model is \( l \). The model hidden size is \( d \). Let \(\mathbf{X}\in \mathbb{R}^{b \times n \times d}\) represents the data embedding, where \( b \) is the batch size and \( n \) is the number of tokens. The input to the \( i \)th layer is denoted as \(\mathbf{X_{\text{in}}^i}\in \mathbb{R}^{b \times n \times d} \), and the output from the \( i \)th layer is denoted as \(\mathbf{X_{\text{out}}^i}\in \mathbb{R}^{b \times n \times d} \). A learnable weight $S[i]$ is assigned to the $i$th transformer layer, and the set of all layer weights is denoted as $S$.

According to Algorithm~\ref{algorithm1}, we initialize the weight of each layer to 1. As shown in Figure~\ref{figure2}, the output of each layer is scaled by its associated weight before being passed as input to the next layer. To learn the weight of each layer, we employ the input data embedding \( \mathbf{X} \). When pruning \( n \) layers, \( \mathbf{X}\) is repeatedly used as input in each iteration. The objective function in Equation~\ref{equation1} comprises two parts: the language modeling loss \( \mathcal{L}(\mathbf{W}, \mathbf{X}) \), and the sum of the $\ell_1$-norm of all layer weights, $\lambda_1$ balances the two components, $l_1$ is the number of layers in the current model that are not masked.
\begin{equation}
\label{equation1}
\begin{split}
\mathcal{L}_{\text{learn}} = \mathcal{L}(\mathbf{W}, \mathbf{X}) +\lambda_1\sum_{i=0}^{l_1-1}\|S[i]\|_1
\end{split}
\end{equation}
Forward and backward propagation is then performed to learn the set of layer weights \( S \). Subsequently, the layer with the smallest weight is identified and masked out in the next iteration, and its index is added to the pruning set $P$. This process is iteratively performed
$n$ times if there are $n$ layers need to be pruned which follows a greedy strategy. We also explored a one-shot pruning approach, in which a single forward and backward propagation is used to identify the \( n \) layers with the lowest weights. However, as shown in Section~\ref{4.4}, this often results in the removal of consecutive transformer layers, which leads to a substantial degradation in model performance.

The process of minimizing the function in Equation~\ref{equation1} is treated as an optimization task. Since $\ell_1$-norm is not differentiable, backpropagation (BP) can't be used directly, we need to transform the problem using Proposition~\ref{l1-norm}. 
\begin{proposition}
\label{l1-norm}
\textnormal{(If the objective function contains an $\ell_1$ regularization term, it can still be optimized using BP. Proof in Appendix \ref{l1})}. The following unconstrained optimization problem is equivalent to the constrained optimization problem, where $\|\cdot\|_1$ denotes the \( \ell_1 \)-norm.
\begin{equation}
\label{equation2}
\begin{aligned}
\min \quad||x||_1  \hspace{0.5em} \iff \hspace{0.5em} \min_{x, y}& \quad\mathbf{1}^T y \\
\text{s.t.}& \quad -y \leq x \leq y, \\
\quad &\quad y \geq 0.
\end{aligned}
\end{equation}
\end{proposition}
The objective function in Equation~\ref{equation1} can be reformulated as an equivalent constrained problem in Equation~\ref{equation3}. After transformation, the objective function is differentiable, and can be solved by BP.
\begin{equation}
\label{equation3}
\begin{aligned}
\min_{W,y}& \quad\mathcal{L}(\mathbf{W}, \mathbf{X}) + \lambda_1 \mathbf{1}^T y  \\
\text{s.t.}& \quad -y \leq S \leq y, \\
\quad &\quad y \geq 0.
\end{aligned}
\end{equation}
\subsection{The Second Stage Regularization}
\begin{figure}[ht]
\begin{center}
\centerline{\includegraphics[width=0.7\columnwidth]{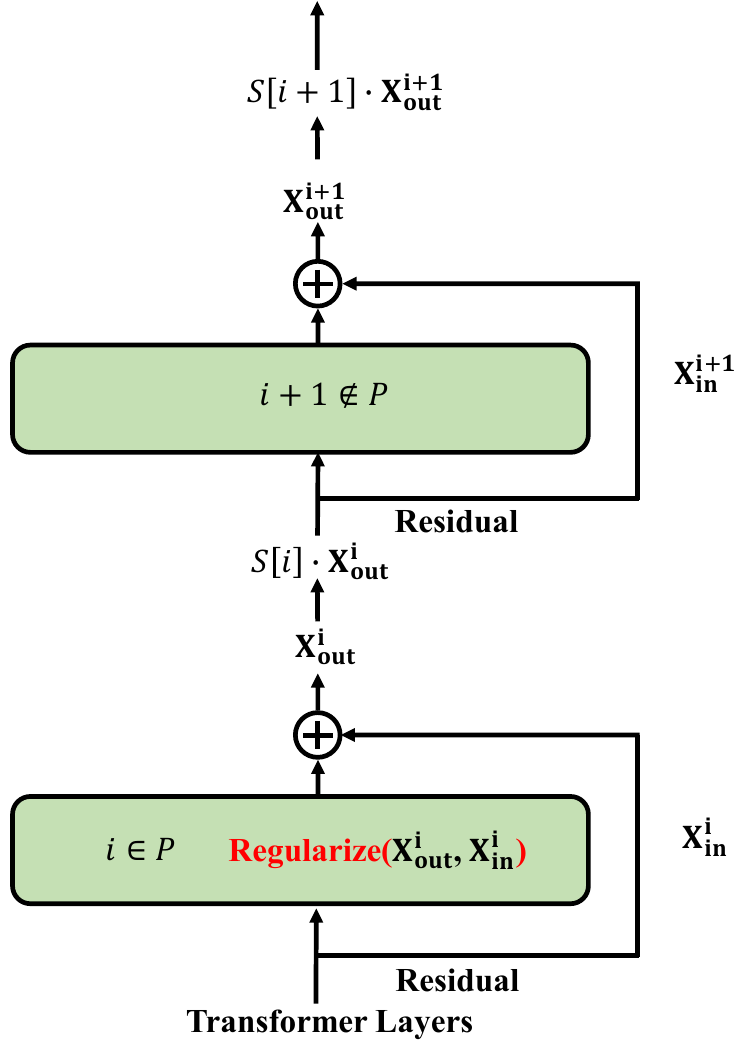}}
\caption{Details of layer weight learning and regularization. $P$ is the set of prunable layers, identified iteratively using the method in Section~\ref{3.2}.}
\label{figure2}
\end{center}
\end{figure}
After learning the layer weights, we further lighten the impact of the pruned layers on the final model output by applying regularization to them. 
The approach is simple and straightforward: we use the data embedding \( \mathbf{X} \) as input and apply a one-shot regularization on the difference between the output and input of each layer in the pruning set \( P \). For example, in Figure~\ref{figure2}, the $i$th layer in \( P \) and should be regularized, we just add the $\ell_1$-norm or $\ell_2$-norm of $\mathbf{X_{out}^i-X_{in}^i}$ to the loss function. This encourages the knowledge to be redistributed from the pruned layers to the remaining ones, thereby reducing the amount of knowledge retained in the pruned layers and significantly minimizing the performance degradation caused by subsequent pruning.

The objective function in Equation~\ref{equation4} comprises two parts: the language modeling loss $\mathcal{L}(\mathbf{W}, \mathbf{X})$ and the regularization loss, $\lambda_2$ balances the two components. When the \( \ell_1 \)-norm is used, the equivalent constrained optimization problem transformed using Proposition~\ref{l1-norm} is shown in Equation~\ref{equation5}. This formulation can be directly solved using BP.
\begin{equation}
\label{equation4}
\begin{split}
\mathcal{L}_{\text{sum}} = \mathcal{L}(\mathbf{W}, \mathbf{X}) +\lambda_2\sum_{i=0}^{|P|-1}\|\mathbf{X_{out}^i-X_{in}^i}\|
\end{split}
\end{equation}
\begin{equation}
\label{equation5}
\begin{aligned}
\min_{W,Y_i}& \quad \mathcal{L}(\mathbf{W}, \mathbf{X}) +\lambda_2\sum_{i=0}^{|P|-1}\mathbf{1}^T \mathbf{Y_i} \mathbf{1} \\
\text{s.t.}&\quad -\mathbf{Y_i} \leq \mathbf{X_{\text{out}}^i - X_{\text{in}}^i} \leq \mathbf{Y_i},  \\
&\quad\mathbf{ Y_i} \geq 0.
\end{aligned}
\end{equation}
\subsection{Pruning}
After applying the regularization, we directly remove the transformer layers in the set \( P \).
\section{Experiments}
\label{section4}
This section introduces experimental setup~(\ref{4.1}) and analyzes the effectiveness of TRSP  from the following aspects: performance comparison~(\ref{4.2}), acceleration~\ref{4.3}, robustness under different pruning ratios~(\ref{4.6}), dependency on different datasets~(\ref{4.7}), low overhead~(\ref{4.8}), and ablation study~(\ref{4.9}), choice of learning layer weights~(\ref{4.4}), impact of regularization~(\ref{4.5}).
\subsection{Experimental Setup}
\label{4.1}
\textbf{Datasets:} We evaluated on generation and zero-shot tasks. For the generation task, following prior work\cite{ashkboos2024slicegpt}, we evaluate the model’s perplexity on WikiText-2~\cite{merity2016pointer} test set. For zero-shot task, we evaluate on PIQA~\cite{bisk2020piqa}, WinoGrande~\cite{sakaguchi2021winogrande}, HellaSwag~\cite{zellers2019hellaswag}, ARC-e and ARC-c~\cite{clark2018think}. In Section~\ref{4.7} we randomly selected data from Alpaca~\cite{alpaca}, WikiText-2~\cite{merity2016pointer}, PTB~\cite{marcus1993building}, and C4~\cite{raffel2020exploring}. 

\textbf{Implementation:} All methods are developed in PyTorch~\cite{paszke2019pytorch}, leveraging the Hugging Face Transformers library~\cite{wolf2019huggingface}. Experimental evaluations are carried out on 80GB 
 NVIDIA A100 GPUs. Pruned models are evaluated with llm-eval-harness~\cite{eval-harness}. More details are in Appendix \ref{B}.

\textbf{Evaluation Metrics:} The ability on generation task is evaluated by \textit{perplexity}, a well-established and robust metric~\cite{yao2022zeroquant}. Zero-shot tasks ability is evaluated using \textit{accuracy}~\cite{dong2024pruner}. Acceleration is measured by \textit{throughput} and \textit{latency}~\cite{song2024sleb}.

\textbf{Models:} The models include the Phi-2~\cite{javaheripi2023phi}, OPT models (OPT-2.7B, OPT-13B)~\cite{zhang2022opt}, and LLaMA models (LLaMA2-7B, LLaMA2-13B, LLaMA3-8B)~\cite{touvron2023llama,grattafiori2024llama}.

\textbf{Baselines:} We compare TRSP with strong layer-wise structured pruning methods: SLEB~\cite{song2024sleb}, ShortGPT~\cite{men2024shortgpt}, LaCo~\cite{yang2024laco}, Shortened LLaMA (PPL version)~\cite{kim2024shortened}.

\subsection{Performance Comparison}
\label{4.2}
To ensure fairness, 128 sequences of length 2048 were randomly drawn from the WikiText-2 training set for TRSP’s two-stage regularization and the calibration process of the baselines. Following \cite{ashkboos2024slicegpt}, we selected another 1,000 samples from WikiText-2 training dataset to retrain leveraging LoRA~\cite{hu2021lora} on the baselines after pruning. The aforementioned 128 samples and the 1,000 samples have no overlap. Because TRSP is retraining-free, we did not retrain TRSP. The pruning ratio was 25\% which means 25\% of the transformer layers in a model will be removed.
\begin{table*}[t]
\centering
\caption{Performance comparison of TRSP and baselines. `PR' is the pruning ratio. `PPL' is the perplexity on WikiText-2. The accuracy is evaluated on five zero-shot benchmarks. TRSP -\(\ell_2\) means using the \(\ell_2\)-norm in the second stage regularization, TRSP -\(\ell_1\) is using the \(\ell_1\)-norm. The best result is in \textbf{bold}, the second-best is \underline{underlined}.}
\label{table1}
\adjustbox{width=1.0\textwidth}{
\begin{tabular}{cccccccccccc}
\toprule
\textbf{Model} & \textbf{Method} & \textbf{PR} & \textbf{PPL ($\downarrow$)} & \textbf{PIQA(\%)} & \textbf{WinoGrande(\%)} & \textbf{HellaSwag(\%)} & \textbf{ARC-e(\%)} & \textbf{ARC-c(\%)} & \textbf{Avg\_Acc(\%)} \\
\midrule
\multirow{6}{*}{\textbf{Phi-2}} 
& Dense       & 0\%  &  5.28 & 79.11 & 75.77 & 73.83 & 78.32 & 54.18 & 72.24\\
& SLEB    & 25\%  & 7.65 &68.85  &63.63  &50.96 & 49.38 & 30.79 &52.72 \\
& ShortGPT    & 25\% &7.15  &69.67 &65.19 &51.26 &\textbf{52.46}  &33.89  &54.49  \\
& LaCo & 25\% &7.38  & 68.53 &63.76 &50.39 &51.28  &33.45  &53.48  \\
& Shortened LLaMA & 25\% &7.74 & 66.88 &62.19 &51.45 & 51.47 &32.66 &52.93  \\
& TRSP -\(\ell_2\) \cellcolor[gray]{0.8}& 25\%\cellcolor[gray]{0.8} & \textbf{6.53} \cellcolor[gray]{0.8}& \underline{71.35} \cellcolor[gray]{0.8}& \textbf{67.62}\cellcolor[gray]{0.8} &  \underline{55.84} \cellcolor[gray]{0.8}& \underline{52.26}\cellcolor[gray]{0.8}& \underline{35.75}\cellcolor[gray]{0.8} & \textbf{56.56} \cellcolor[gray]{0.8}\\
& TRSP -\(\ell_1\)\cellcolor[gray]{0.8}& 25\%\cellcolor[gray]{0.8} & \underline{6.58} \cellcolor[gray]{0.8}& \textbf{71.42} \cellcolor[gray]{0.8}& \underline{67.13}\cellcolor[gray]{0.8} & \textbf{56.05} \cellcolor[gray]{0.8}& 52.18\cellcolor[gray]{0.8}& \textbf{35.79}\cellcolor[gray]{0.8} & \underline{56.51} \cellcolor[gray]{0.8}\\
 \midrule
\multirow{6}{*}{\textbf{OPT-2.7B}} 
& Dense       & 0\%  &12.46  & 74.81 & 61.01 & 60.58 & 54.42 &31.14 & 56.39 \\
& SLEB    & 25\%  & 15.71 & 65.19 &56.26 & 44.54& 46.28 & 25.14 & 47.48 \\
& ShortGPT & 25\%  &14.96  & 67.37 &57.65 &\textbf{46.82} &49.43  &26.54  &49.56  \\
& LaCo & 25\% &15.38  & 66.54 &59.45 &43.68 & 48.74 &24.96  &48.67  \\
& Shortened LLaMA &25\% &15.89 & 63.14 &57.36&43.57 &47.62  &25.88  &47.51 \\
& TRSP -\(\ell_2\) \cellcolor[gray]{0.8}& 25\%\cellcolor[gray]{0.8} & \underline{13.18} \cellcolor[gray]{0.8}& \underline{70.54} \cellcolor[gray]{0.8}& \textbf{60.27}\cellcolor[gray]{0.8} & \underline{46.35} \cellcolor[gray]{0.8}& \underline{51.59}\cellcolor[gray]{0.8}& \underline{27.36}\cellcolor[gray]{0.8} & \underline{51.22} \cellcolor[gray]{0.8}\\
& TRSP -\(\ell_1\)\cellcolor[gray]{0.8}& 25\%\cellcolor[gray]{0.8} & \textbf{13.12} \cellcolor[gray]{0.8}& \textbf{70.65} \cellcolor[gray]{0.8}& \underline{60.13}\cellcolor[gray]{0.8} & 46.24 \cellcolor[gray]{0.8}& \textbf{51.76}\cellcolor[gray]{0.8}& \textbf{27.55}\cellcolor[gray]{0.8} & \textbf{51.27} \cellcolor[gray]{0.8}\\
\midrule
\multirow{6}{*}{\textbf{LLaMA2-7B}} 
 & Dense        & 0\%  & 5.47 & 79.11 &69.06 & 75.99 & 74.58 & 46.25 & 69.00 \\
& SLEB    & 25\%  & 9.63 & 65.22 & 63.38 & 55.51 & 56.39 & 33.46& 54.79 \\
& ShortGPT    & 25\%   &8.89 &66.75 &66.26 &57.14 &58.93  &36.42  &57.10  \\
& LaCo          & 25\% &9.14 &69.45 & 65.31&52.67 &55.73  &34.89  &55.61  \\
& Shortened LLaMA & 25\% &9.47 & 65.58 &64.72 &58.36 &54.19  &32.96  &55.16  \\
 &   TRSP -\(\ell_2\) \cellcolor[gray]{0.8}& 25\%\cellcolor[gray]{0.8} & \textbf{7.08} \cellcolor[gray]{0.8}& \textbf{72.48} \cellcolor[gray]{0.8}& \underline{67.52}\cellcolor[gray]{0.8} & \textbf{60.45} \cellcolor[gray]{0.8}& \textbf{62.69}\cellcolor[gray]{0.8}& \underline{39.73}\cellcolor[gray]{0.8} & \textbf{60.57} \cellcolor[gray]{0.8}\\
 &   TRSP -\(\ell_1\)\cellcolor[gray]{0.8}& 25\%\cellcolor[gray]{0.8} & \underline{7.17} \cellcolor[gray]{0.8}& \underline{72.08} \cellcolor[gray]{0.8}& \textbf{67.93}\cellcolor[gray]{0.8} & \underline{60.42} \cellcolor[gray]{0.8}& \underline{62.38}\cellcolor[gray]{0.8}& \textbf{39.89}\cellcolor[gray]{0.8} & \underline{60.54} \cellcolor[gray]{0.8}\\
\midrule
\multirow{6}{*}{\textbf{LLaMA3-8B}} 
 & Dense  & 0\%   & 5.76  & 85.56 & 77.94 & 79.27 & 78.84 &56.49 & 75.62 \\
& SLEB    & 25\%  & 10.38  & 72.74 & 64.12 & 67.74 & 65.84 &45.16 &63.12 \\
& ShortGPT& 25\%  &9.26  & 75.38 &69.25 &70.12 &68.54  &47.57  &66.17  \\
& LaCo    &25\%   &10.14 & 74.69 &67.52 &66.36 &69.43  &46.31  &64.86  \\
& Shortened LLaMA & 25\% &9.84&73.62 &68.73 &68.55 &66.16  &43.39  &64.09  \\
 &   TRSP -\(\ell_2\) \cellcolor[gray]{0.8}& 25\%\cellcolor[gray]{0.8} &\underline{7.84}\cellcolor[gray]{0.8}&\underline{77.25}\cellcolor[gray]{0.8}& \textbf{71.63}\cellcolor[gray]{0.8} & \underline{72.26}\cellcolor[gray]{0.8}& \textbf{71.49}\cellcolor[gray]{0.8}& \underline{49.58}\cellcolor[gray]{0.8} & \textbf{68.44}\cellcolor[gray]{0.8}\\
 &   TRSP -\(\ell_1\)\cellcolor[gray]{0.8}& 25\%\cellcolor[gray]{0.8} &\textbf{7.68} \cellcolor[gray]{0.8}&\textbf{77.36} \cellcolor[gray]{0.8}& \underline{71.33}\cellcolor[gray]{0.8} &\textbf{72.82} \cellcolor[gray]{0.8}& \underline{70.75}\cellcolor[gray]{0.8}& \textbf{49.66}\cellcolor[gray]{0.8} & \underline{68.38}\cellcolor[gray]{0.8}\\
\midrule
\multirow{6}{*}{\textbf{OPT-13B}} 
 & Dense       & 0\%  &10.12 & 76.82 & 64.80& 69.81 & 61.87 & 35.67 &61.79 \\
& SLEB    & 25\%  & 11.96 & 72.83 & 64.06 & 63.32 & 59.98 & 34.65 & 58.97 \\
& ShortGPT    & 25\%  &11.38&73.59 &\textbf{64.52} &65.68 &60.41  &\underline{34.99}  &59.84  \\
& LaCo & 25\%   &11.79 &73.96 &63.24 &62.17  &\underline{61.46} &33.28 & 58.82 \\
& Shortened LLaMA & 25\% &11.62&71.26& 63.57& 66.29& 58.47 &33.83 &58.68  \\
 & TRSP -\(\ell_2\) \cellcolor[gray]{0.8}& 25\%\cellcolor[gray]{0.8} & \underline{10.45} \cellcolor[gray]{0.8}& \underline{74.15} \cellcolor[gray]{0.8}& \underline{64.47}\cellcolor[gray]{0.8} & \textbf{68.82} \cellcolor[gray]{0.8}& \textbf{61.55}\cellcolor[gray]{0.8}& \textbf{35.23}\cellcolor[gray]{0.8} &\textbf{60.84} \cellcolor[gray]{0.8}\\
 & TRSP -\(\ell_1\)\cellcolor[gray]{0.8}& 25\%\cellcolor[gray]{0.8} &\textbf{10.32} \cellcolor[gray]{0.8}& \textbf{74.58} \cellcolor[gray]{0.8}& 64.37\cellcolor[gray]{0.8} & \underline{68.45} \cellcolor[gray]{0.8}& 61.29\cellcolor[gray]{0.8}& 34.87\cellcolor[gray]{0.8} & \underline{60.71} \cellcolor[gray]{0.8}\\
 \midrule
\multirow{6}{*}{\textbf{LLaMA2-13B}} 
 & Dense  & 0\%  & 4.88 & 80.47 & 72.22 & 79.39 &77.48 &49.23 & 71.76 \\
& SLEB    & 25\%   & 7.08 & 68.31 & 66.86 & 57.12 & 62.19 & 38.45 & 58.59 \\
 & ShortGPT    & 25\% &6.79  &73.26 &68.37 &62.25 &67.54  &43.38  &62.96  \\
& LaCo & 25\%       &7.14  & 72.35 &65.78 &59.43 &65.36  &42.73  &61.13  \\
& Shortened LLaMA & 25\% &6.92  & 69.45 &66.38 &59.86& 65.25 &39.12  &60.01  \\
 &   TRSP -\(\ell_2\) \cellcolor[gray]{0.8}& 25\%\cellcolor[gray]{0.8} &\textbf{5.82}  \cellcolor[gray]{0.8}& \underline{74.56} \cellcolor[gray]{0.8}& \textbf{69.34}\cellcolor[gray]{0.8} & \underline{64.79} \cellcolor[gray]{0.8}& \underline{71.25}\cellcolor[gray]{0.8}& \textbf{45.63}\cellcolor[gray]{0.8} & \underline{65.11} \cellcolor[gray]{0.8}\\
 &   TRSP -\(\ell_1\)\cellcolor[gray]{0.8}& 25\%\cellcolor[gray]{0.8} & \underline{5.89}\cellcolor[gray]{0.8}& \textbf{75.06} \cellcolor[gray]{0.8}& \underline{69.18}\cellcolor[gray]{0.8} & \textbf{64.96} \cellcolor[gray]{0.8}& \textbf{71.43}\cellcolor[gray]{0.8}& \underline{45.26}\cellcolor[gray]{0.8} & \textbf{65.18} \cellcolor[gray]{0.8}\\
\bottomrule
\end{tabular}}
\end{table*}
As shown in Table~\ref{table1}, TRSP achieves the lowest perplexity and the highest average accuracy across all models, demonstrating its superior performance on both \textit{generation} and \textit{zero-shot} tasks. Notably, on OPT-13B, TRSP only drops 1\% in average accuracy compared to the dense model. On LLaMA2-7B, its perplexity is 20\% lower than the second-best method, ShortGPT. This further demonstrates the effectiveness of TRSP. The minimal performance difference between \(\ell_2\)-norm and \(\ell_1\)-norm suggests that TRSP is not sensitive to the choice of regularization norm. In the following sections, we refer to using the \(\ell_2\)-norm in the stage two regularization.

\subsection{Acceleration}
\label{4.3}
LLMs language processing involves two key phases with different bottlenecks: compute-bound prompt processing and memory-bound token generation. We measured the speedup for each stage individually. Table~\ref{table2} presents the latency and throughput results for OPT-13B and LLaMA2-13B running on a single 80GB NVIDIA A100 GPU. Following prior methods ~\cite{song2024sleb}, the token generation was tested by producing 128-token sentences with a batch size of 64, and prompt processing latency was assessed with a 2048-token input.
\begin{table*}[ht]
\centering
\caption{Throughput (tokens/s) and latency (ms) on OPT-13B and LLaMA2-13B. `PPL’ is the perplexity on Wikitext2. `PR' is pruning ratio. `TI' is throughput increase.}
\label{table2}
\adjustbox{width=0.8\textwidth}{
\begin{tabular}{cccccccccccccc}
\toprule
\textbf{Model} & \textbf{Method} & \textbf{PR}&\textbf{PPL($\downarrow$)} &\textbf{Avg\_Acc(\%)}& \textbf{Tokens/s($\uparrow$)} & \textbf{TI($\uparrow$)} & \textbf{Latency($\downarrow$)} & \textbf{Speedup($\uparrow$)}  \\
\midrule
\multirow{3}{*}{\textbf{OPT-13B}} 
 & Dense       & 0\% &10.12 &61.79&1029 & 1.00× & 386.5& 1.00×  \\
 & TRSP \cellcolor[gray]{0.8}& 25\%\cellcolor[gray]{0.8} & 10.45\cellcolor[gray]{0.8} &60.84\cellcolor[gray]{0.8}&1348\cellcolor[gray]{0.8}&1.31×\cellcolor[gray]{0.8}& 286.3 \cellcolor[gray]{0.8}& 1.35×\cellcolor[gray]{0.8}  \\
  & TRSP  \cellcolor[gray]{0.8}& 50\%\cellcolor[gray]{0.8}  &15.38\cellcolor[gray]{0.8}&50.83\cellcolor[gray]{0.8}& 1801\cellcolor[gray]{0.8}&1.75×\cellcolor[gray]{0.8}& 208.9 \cellcolor[gray]{0.8}& 1.85×\cellcolor[gray]{0.8}  \\
 \midrule
\multirow{3}{*}{\textbf{LLaMA2-13B}} 
 & Dense        & 0\% & 4.88&71.76&1066 & 1.00× & 396.9 & 1.00×  \\
 & TRSP  \cellcolor[gray]{0.8}& 25\%\cellcolor[gray]{0.8} &5.82\cellcolor[gray]{0.8} &65.11\cellcolor[gray]{0.8}&1386 \cellcolor[gray]{0.8}& 1.30×\cellcolor[gray]{0.8}&298.4\cellcolor[gray]{0.8} & 1.33× \cellcolor[gray]{0.8}\\
  & TRSP  \cellcolor[gray]{0.8}& 50\%\cellcolor[gray]{0.8} &11.28\cellcolor[gray]{0.8} &57.57\cellcolor[gray]{0.8}&1823 \cellcolor[gray]{0.8}& 1.71×\cellcolor[gray]{0.8}& 216.9\cellcolor[gray]{0.8} & 1.83× \cellcolor[gray]{0.8}\\
\bottomrule
\end{tabular}}
\end{table*}

Pruning OPT-13B by 50\% with TRSP yields a 75\% increase in throughput and a 46\% reduction in latency compared to the dense model. For LLaMA2-13B, it delivers a 71\% improvement in throughput and a 45\% decrease in latency. These results underscore the end-to-end acceleration achieved by TRSP.
\subsection{Robustness to Different Pruning Ratios}
\label{4.6}
\begin{figure}[ht]
\begin{center}
\centerline{\includegraphics[width=\columnwidth]{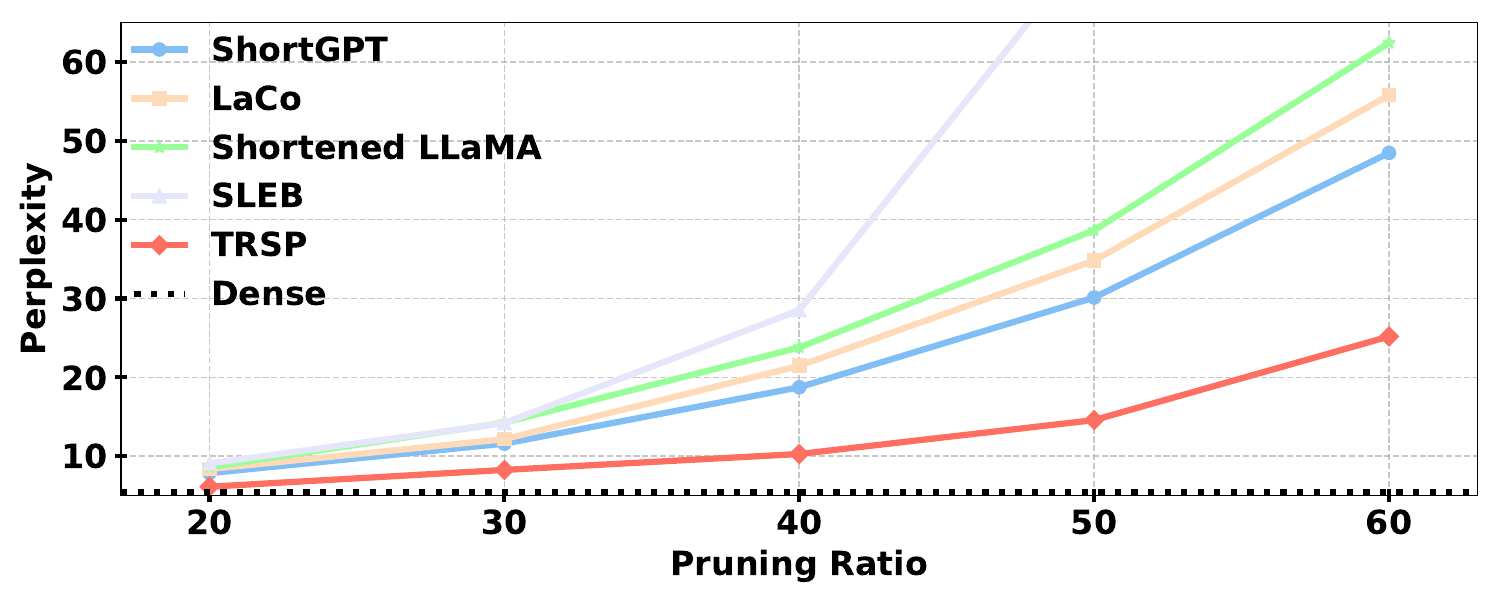}}
\caption{Perplexity of LLaMA2-7B pruned by five methods under different pruning ratios on WikiText-2.}
\label{figure5}
\end{center}
\end{figure}

Using the same settings as Section~\ref{4.2}, we vary the pruning ratio from 20\% to 60\%. As shown in Figure~\ref{figure5}, TRSP consistently achieves lower perplexity than other methods. At 20\% pruning, it matches the dense model, and even at 60\%, where SLEB fails, it maintains low perplexity. This demonstrates TRSP's robustness and effectiveness in structured pruning for model acceleration. Additional results are provided in Appendix~\ref{E}.
\subsection{Dependency on Datasets}
\label{4.7}
\begin{figure}[htbp]
\begin{center}
\centerline{\includegraphics[width=\columnwidth]{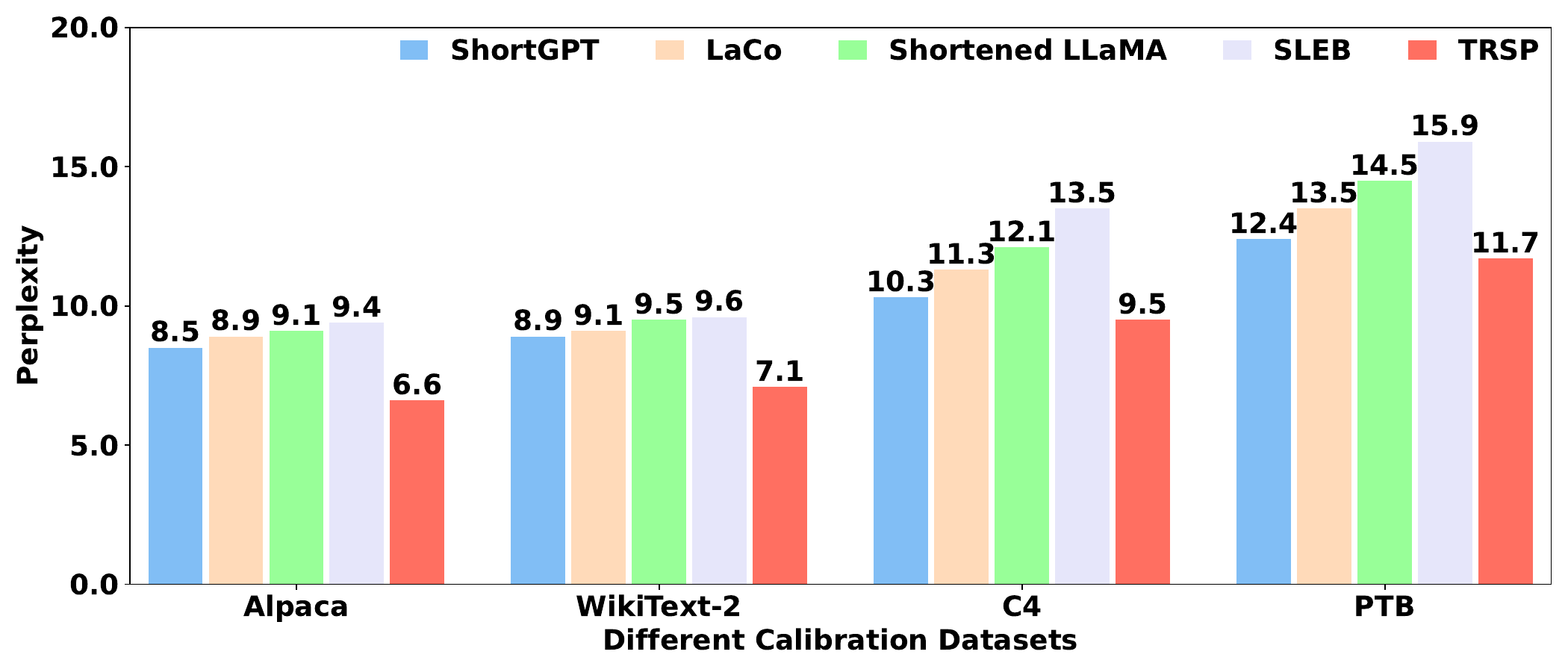}}
\caption{Comparison of perplexity on different calibration datasets at a pruning ratio of 25\% on LLaMA2-7B.}
\label{figure6}
\end{center}
\end{figure}

Since TRSP relies on data-driven regularization, we investigate its dataset dependency. Keeping all other settings consistent with Section \ref{4.2}, we only change the source of the 128 samples: they are drawn respectively from Alpaca, WikiText-2, PTB, and C4. We evaluated perplexity of five methods on WikiText-2. As shown in Figure~\ref{figure6}, TRSP consistently outperforms the other methods across datasets, demonstrating its robustness.
\subsection{Minimal Cost}
\label{4.8}
\begin{figure}[htbp]
\begin{center}
\centerline{\includegraphics[width=\columnwidth]{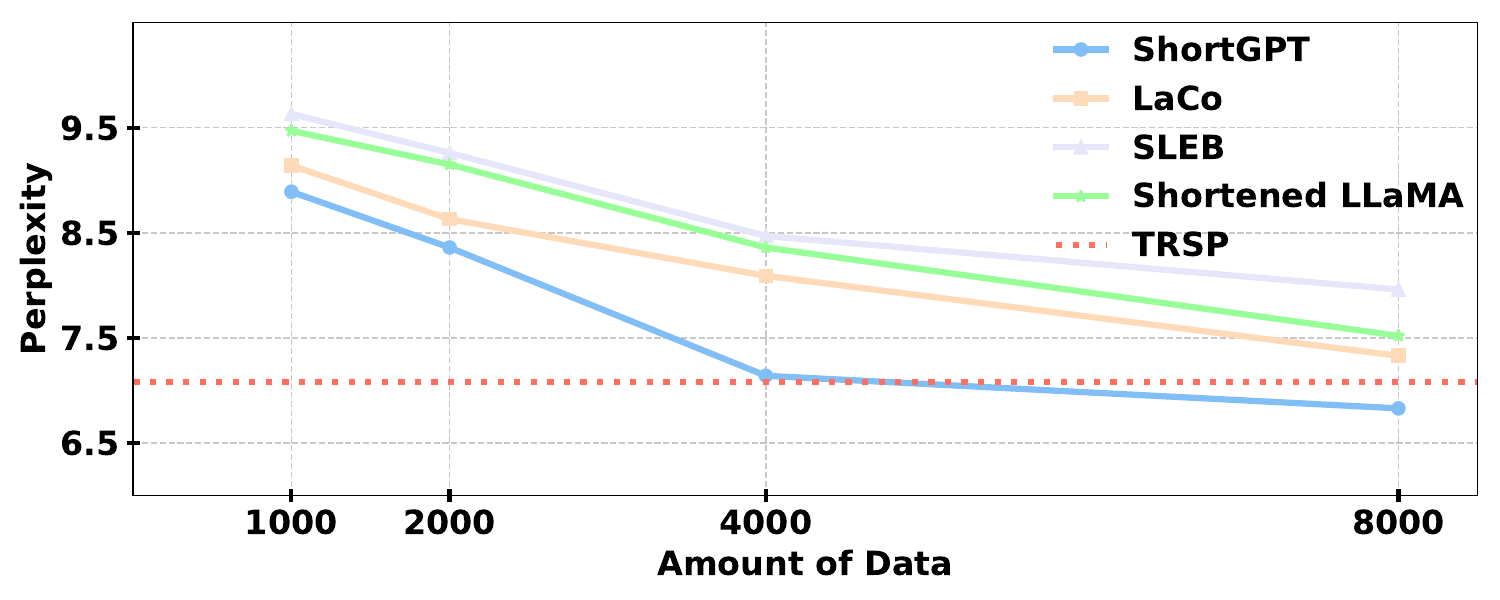}}
\caption{The perplexity on WikiText-2 using TRSP and baselines under different amounts of retraining data.}
\label{figure7}
\end{center}
\end{figure}

We only vary the retraining data size from 1,000 to 8,000 and use the same settings as in Section~\ref{4.2}. We evaluated the perplexity of baselines on LLaMA2-7B. TRSP is retraining-free. From Figure \ref{figure7}, we observe that TRSP outperforms the other methods with 4,000 retraining data, significantly reducing the cost. We speculate that TRSP iteratively learns layer weights and then applies regularization, allowing it to identify layers to prune more accurately, and then transfer the knowledge from those layers to the remaining layers of the model through regularization. This process reduces knowledge loss, thus preserving model performance and lowering the retraining cost.

\subsection{Ablation Study}
\begin{table}[ht]
\centering
\captionof{table}{Ablation results on LLaMA2-7B and LLaMA2-13B. `w/o W' denotes learning layer weights in one-shot, `w/o R' means no regularization. The pruning ratio is 25\%.}
\label{table6}
\adjustbox{width=\columnwidth}{
\begin{tabular}{cccccc}
\toprule
\textbf{Model} & \textbf{Setting} & \textbf{PPL($\downarrow$)} & \textbf{\textbf{$\Delta$}} & \textbf{AVG\_ACC} & \textbf{$\Delta$($\downarrow$)}   \\
\midrule
\multirow{4}{*}{\textbf{LLaMA2-7B}} 
&TRSP  & 7.08 & 0.00 & 60.57 & 0.00 \\
&w/o W & 9.26 & +2.18 & 56.19 & -4.38 \\
&w/o R & 10.15 & +3.07 & 54.36 & -6.21 \\
 \midrule
\multirow{4}{*}{\textbf{LLaMA2-13B}} 
&TRSP  & 5.82  & 0.00 & 65.11 & 0.00 \\
&w/o W & 8.35 & +2.53 & 59.36 & -5.75 \\
&w/o R & 9.47 & +3.65 & 56.25 & -8.86 \\
\bottomrule
\end{tabular}}
\end{table}
\label{4.9}
\textbf{Effect of Learning Layer Weights Iteratively} TRSP  learns layer weights in a greedy and iterative manner. As shown in Table~\ref{table6} (Row 3 and Row 6), replacing iterative layer weight learning with a one-shot approach leads to increased model perplexity and decreased accuracy, highlighting the importance of learning layer weights iteratively, which will be discussed in detail in Section~\ref{4.4}.

\textbf{Effect of Applying Regularization}
As shown in Table~\ref{table6} (Row 4 and Row 7), removing the regularization process results in increased model perplexity and decreased accuracy, demonstrating the effectiveness of applying regularization, which will be discussed in detail in Section~\ref{4.5}.

With all other settings the same as Section~\ref{4.2}, we evaluate the model's performance by varying \(\lambda_1 \in [10^{-5}, 10^{-4}, 10^{-3}, 5 \times 10^{-3}, 10^{-2}, 5 \times 10^{-2}, 10^{-1}]\) and \(\lambda_2 \in [10^{-5}, 10^{-4}, 10^{-3}, 5 \times 10^{-3}, 10^{-2}, 5 \times 10^{-2}, 10^{-1}]\). We perform a grid search over \(\lambda_1\) and \(\lambda_2\), with details provided in Appendix~\ref{C}. The optimal combination yielding the lowest perplexity on LLaMA2-7B is \(\lambda_1 = 5 \times 10^{-3}\) and \(\lambda_2 = 10^{-3}\).
\subsection{Choice of Learning Layer Weights}
\label{4.4}
In this section, we explore the effectiveness of (1) iteratively learning layer weights using a greedy strategy compared to (2) acquiring all layer weights at one-shot. We set the pruning ratio to 25\% and conduct experiments on LLaMA2-7B (32 layers), using the same settings as in Section~\ref{4.2}. 

\begin{figure}[ht]
\begin{center}
\centerline{\includegraphics[width=\columnwidth]{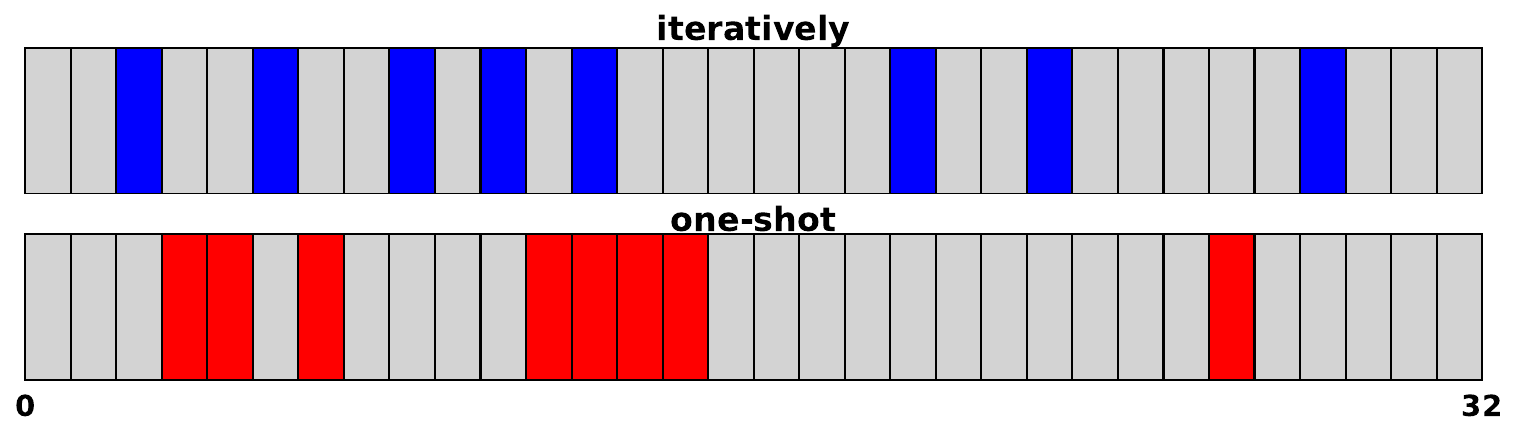}}
\caption{Results of selecting the eight lowest-weight layers using iterative and one-shot layer weight learning.}
\label{figure3}
\end{center}
\end{figure}

The perplexity and average accuracy of the pruned models are shown in Table~\ref{table3}. It can be observed that, compared to using a greedy strategy to iteratively learn layer weights, learning all layer weights in one-shot exhibits significant degradation in model performance. This behavior can be explained by the observation that the importance of a block changes as other blocks are removed. The results of selecting the eight least important layers using both methods are shown in Figure~\ref{figure3}. It can be seen that learning layer weights in one-shot tends to select consecutive layers. While these blocks may individually have limited impact on LLM inference performance, removing a continuous sequence of blocks can significantly degrade the overall inference results.
\begin{table}[ht]
\centering
\captionof{table}{The pruned LLaMA2-7B performance under different learning layer weights methods.}
\label{table3}
\adjustbox{width=0.5\columnwidth}{
\begin{tabular}{ccccccccc}
\toprule 
\textbf{Cases} &\textbf{PPL($\downarrow$)} & \textbf{Avg\_Acc(\%)}\\
\midrule
\textbf{(1)} &7.08& 60.57 \\
\midrule
\textbf{(2)} &9.26 & 56.19  \\
\bottomrule
\end{tabular}}
\end{table}
\vspace{-15pt}
\vspace{-20pt}
\subsection{Impact of Stage Two Regularization}
\label{4.5}
We keep other settings consistent with Section~\ref{4.2}. After learning the layer weights iteratively, we consider two scenarios: (1) without regularization and (2) with regularization, then prune. The perplexity and average accuracy under (1) and (2) on LLaMA2-7B are in Table~\ref{table4}. The model exhibits lower perplexity and higher average accuracy with regularization. Since perplexity is the exponential form of cross-entropy loss, lower perplexity corresponds to a lower cross-entropy loss. A smaller loss function indicates better model performance indicating that the regularization process mitigates the impact of pruning on the overall model.

According to previous work \cite{liu2023deja}, the output and input of each layer in LLMs exhibit high similarity. This inherent similarity enables us to apply regularization on the difference between the input and output of certain layers using only a small portion of data. We compute the cosine similarity according to Equation~\ref{equation6} on the LLaMA2-7B model between the input representations \( \mathbf{X_{\text{in}}^i} \in \mathbb{R}^{b \times n \times d} \) and the output representations \( \mathbf{X_{\text{out}}^i}\in \mathbb{R}^{b \times n \times d} \), where \( b \) denotes the batch size, \( n \) the number of tokens, and \( d \) is the hidden size of the model. The computation is performed on layers [3, 6, 9, 11, 13, 20, 23, 29], which are selected using the iterative method described in Section~\ref{4.4}, both before and after regularization.

\begin{figure}[htbp]
\begin{center}
\centerline{\includegraphics[width=\columnwidth]{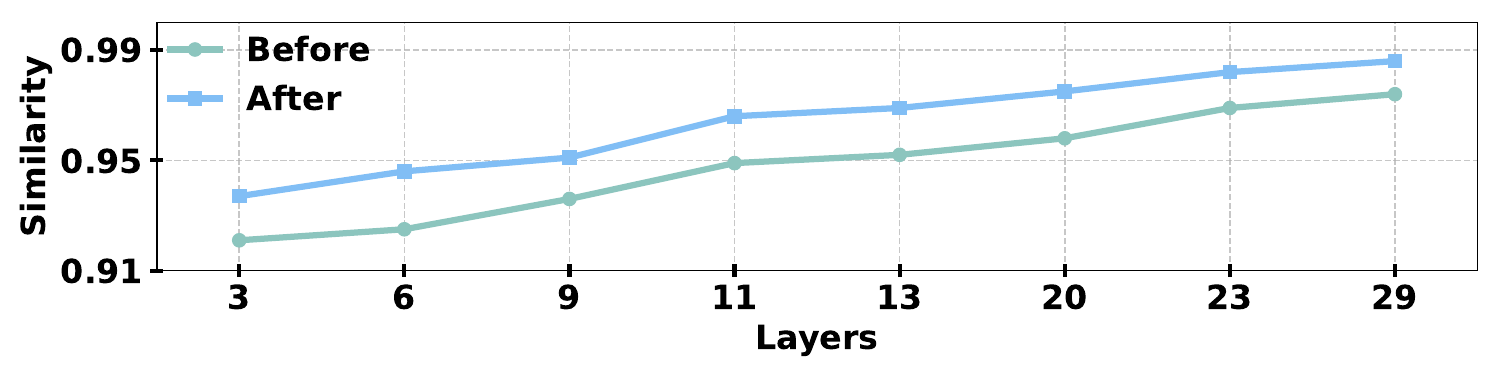}}
\caption{The input-output similarity of the regularized layers.}
\label{figure4}
\end{center}
\end{figure}

As shown in Figure \ref{figure4}, the input-output similarity of these layers is already high before regularization and increases even further after regularization. The increased similarity in the layers with regularization indicates that the input undergoes less change after passing through these layers, suggesting that less knowledge is retained in them. In the Appendix~\ref{F}, we illustrate the changes in similarity between the input and output of the layers that were not regularized, before and after regularization. The results in Figure~\ref{figure8} of Appendix~\ref{F} show that the similarity in these layers decreases, meaning that, after regularization, the input undergoes greater changes when passing through these layers than before. This demonstrates that the regularization process weakens the influence of the regularized layers and enhances the influence of the unregularized parts, suggesting that regularization process may facilitate the transfer of knowledge from the regularized layers to the rest of the model.
\vspace{-10pt}
\begin{table}[htbp]
\centering
\captionof{table}{The performance differences of (1) and (2).}
\label{table4}
\adjustbox{width=0.5\linewidth}{
\begin{tabular}{ccc}
\toprule 
\textbf{Cases} & \textbf{PPL($\downarrow$)} & \textbf{Avg\_Acc(\%)}\\
\midrule
\textbf{(1)} & 7.08 & 60.57 \\
\midrule
\textbf{(2)} & 10.15 & 54.36 \\
\bottomrule
\end{tabular}}
\end{table}
\vspace{-10pt}

\section{Conclusion}
\label{5}
We propose a novel structured pruning method, TRSP. By performing two-stage regularization, TRSP retains more knowledge and better preserves model performance compared to direct parameter elimination. TRSP surpasses existing layer-wise pruning methods in generation and zero-shot tasks. For example, it reduces perplexity by 20\% compared to ShortGPT on LLaMA2-7B, under 25\% sparsity, the average accuracy decreases by just 1\%, while delivering a 1.35× acceleration over the dense model. TRSP is retraining-free, significantly lowering computational overhead and dependence on retraining. The novel structured pruning method offers potential guidance for pruning strategies in LLMs.

\section*{Limitations}
TRSP has primarily been evaluated on autoregressive language models and its applicability to other
architectures or tasks remains unexplored. In future work, we plan to explore the application of TRSP to other model architectures, such as convolutional neural networks (CNNs).

\section*{Acknowledgments}
This work is supported by the National Natural Science Foundation of China (No. U21B2010 and 62322120), and the China Postdoctoral Science Foundation (Grant No. 2025T180461 and 2025M771685).


\bibliography{custom}

\appendix
\section{Proof of Proposition \ref{l1-norm}}
\label{l1}

\textbf{Step 1: Expressing \(\ell_1\)-norm Using Elements.} \\
The objective function in the unconstrained problem is the \( \ell_1 \)-norm of the vector \( x \), which is defined as:
\begin{equation}
||x||_1 = \sum_{i=1}^{n} |x_i|
\end{equation}

This function aims to minimize the sum of the absolute values of the components of \( x \).

\textbf{Step 2: Reformulating the Constrained Problem} \\
The constrained optimization problem introduces an auxiliary variable \( y \), where for each element \( i \):
\begin{equation}
x_i \geq -y_i \quad \text{and} \quad   x_i \leq y_i\quad 
\end{equation}
This implies that \( y_i \geq |x_i| \), meaning each element of \( y \) serves as an upper bound for the absolute value of the corresponding element in \( x \). Consequently, minimizing \(|x|_1\) is equivalent to minimizing the sum of the elements in \( y \). Thus, the objective function is defined as:
\begin{equation}
\mathbf{1}^T y
\end{equation}
Thus, minimizing \( \mathbf{1}^T y \) is equivalent to minimizing the sum of the absolute values of \( x \), which is the \( \ell_1 \)-norm of \( x \).

This transformation allows the optimization problem to be solved without directly involving the absolute value function, resulting in an equivalent constrained optimization problem that can be addressed via backpropagation. Thus, the proof of the Proposition~\ref{l1-norm} is complete.

\section{Detailed Implementation}
\label{B}
In this part, we first introduce several hyperparameter settings, with the detailed results shown in Table \ref{table9}. In our experiments, we employ FP16 precision for all evaluated models, including Phi-2, OPT-2.7B, LLaMA3-8B, OPT-13B, LLaMA2-7B, and LLaMA2-13B. For all retraining configurations, we set the LoRA rank \( r \) to 32, the scaling factor \( \alpha \) to 10, and the sequence length to 2048. All other hyperparameters follow the default settings provided in the Hugging Face PEFT package \cite{mangrulkar2022peft}. We set the batch size to 64. In future work, we will further explore a broader range of batch sizes.
\begin{table*}[ht]
\centering
\caption{Implementation Details}
\label{table9}
\adjustbox{width=\textwidth}{
\begin{tabular}{ccccccccccccc}
\toprule
\textbf{Precision} & \textbf{LoRA Rank} & \textbf{Scaling Factor}& \textbf{Max Sequence Length} & \textbf{Batch Size} & \textbf{Learning Rate} & \textbf{Early Stop Threshold} & \textbf{Min Delta}  \\
\midrule 
FP16 & 32       & 10 &2048 &64 & 2e-5 & 5& 1e-4  \\
\bottomrule
\end{tabular}}
\end{table*}
To ensure a fair comparison between TRSP and other methods, we maintain consistency in the data used across all approaches. Specifically, the data used by TRSP for learning layer weights and performing the second stage regularization is identical to the calibration data used by the baseline methods. Furthermore, we ensure that the data employed during the retraining process is consistent across all baseline methods. Following previous works~\cite{song2024sleb}, for the comparison unstructured pruning methods like Magnitude, Wanda, and SparseGPT, we ensure that the data used to compute the importance of individual weights is the same as the data used by TRSP for learning weights and regularization.

\section{Optimal $\lambda_1$ and  $\lambda_2$ for LLaMA2-7B}
\label{C}
Keeping all other settings consistent with Section~\ref{4.2}, we evaluate the model's perplexity on WikiText-2 by varying \(\lambda_1 \in [10^{-5}, 10^{-4}, 10^{-3}, 5 \times 10^{-3}, 10^{-2}, 5 \times 10^{-2}, 10^{-1}]\) and \(\lambda_2 \in [10^{-5}, 10^{-4}, 10^{-3}, 5 \times 10^{-3}, 10^{-2}, 5 \times 10^{-2}, 10^{-1}]\), resulting in a total of 49 combinations. 

It can be observed from Table~\ref{table10} that the optimal combination yielding the lowest perplexity on LLaMA2-7B is \(\lambda_1 = 5 \times 10^{-3}\) and \(\lambda_2 = 10^{-3}\). 

When fixing \(\lambda_2 = 10^{-3}\), we gradually decrease \(\lambda_1\) from \(5 \times 10^{-3}\) to \(10^{-5}\), during which the model's perplexity increases steadily. This suggests that when the \(\ell_1\)-norm loss constitutes a relatively small portion of the total loss during the iterative learning of layer weights, it fails to effectively constrain the layer weights, resulting in larger deviations. Conversely, when \(\lambda_1\) is gradually increased from \(5 \times 10^{-3}\) to \(10^{-1}\), the model's perplexity also increases, indicating that a dominant \(\ell_1\)-norm loss in the total objective function can hinder the optimization of the language modeling capability.

As shown in Table~\ref{table10}, when \(\lambda_1 = 5 \times 10^{-3}\) is fixed, setting \(\lambda_2 = 10^{-5}\) results in a regularization loss that is too weak to effectively redistribute important information. This leads to a substantial increase in perplexity after pruning. On the other hand, when \(\lambda_2 = 10^{-1}\), the overly strong regularization impairs the model’s language modeling capability, also resulting in a noticeable drop in performance after pruning. The best performance is observed at \(\lambda_2 = 10^{-3}\), highlighting the critical need to balance the language modeling loss and regularization loss.

\begin{table*}[ht]
\centering
\caption{The optimal \(\lambda_1\) and \(\lambda_2\) for LLaMA2-7B.}
\label{table10}
\adjustbox{width=0.7\textwidth}{
\begin{tabular}{cccccccc}
\toprule 
$\lambda_1$ / $\lambda_2$&$10^{-5}$ & $10^{-4}$ & $10^{-3}$ & $5 \times 10^{-3}$ & $10^{-2}$& $5 \times 10^{-2}$&$10^{-1}$\\
\midrule
\textbf{$10^{-5}$}        &12.28&11.91&8.94&10.48&13.79&14.73&15.87\\
\textbf{$10^{-4}$}        &11.45&10.36&8.26&9.82&11.87&12.56&13.62\\
\textbf{$10^{-3}$}        &10.82&9.14&7.92&8.53&9.75&10.82&12.98\\
\textbf{$5\times10^{-3}$} 
         
    &10.09&8.35&\textbf{7.08}&7.46             &8.52&10.17&12.25\\
\textbf{$10^{-2}$}        &11.26&9.97&8.13&8.92&10.46&12.88&13.43\\
\textbf{$5\times10^{-2}$} &12.89&11.25&10.38&11.76&12.85&14.75&15.37\\
\textbf{$10^{-1}$}        &14.55&13.58&12.25&14.41&15.72&16.91&17.93\\
\bottomrule
\end{tabular}}
\end{table*}

\section{Performance of TRSP under Different Pruning Ratios and Datasets}
\label{E}
\subsection{The Perplexity of TRSP under Different Pruning Ratios and Datasets}
To systematically evaluate the performance of TRSP on zero-shot tasks across different pruning rates, we adopt the experimental setup outlined in Section \ref{4.2}, in which 128 samples are randomly sampled from the WikiText-2 training set to guide both the iterative learning of layer weights and the regularization process. Subsequently, we evaluate multiple large language models (LLMs) by measuring changes in perplexity across various generative task datasets, including WikiText-2, Alpaca, PTB, and C4, under pruning rates of 10\%, 20\%, 30\%, 40\%, 50\%, and 60\%. The detailed results, presented in Table \ref{table12}, indicate that TRSP exhibits greater robustness as model scale increases, suggesting that the proposed method effectively mitigates performance degradation in larger architectures. This highlights the scalability of TRSP and its potential to maintain model efficiency under varying levels of sparsity.
\begin{table*}[!htbp]
\centering
\caption{Perplexity comparison of TRSP with different pruning ratios. We set the pruning ratios to 10\%, 20\%, 30\%, 40\%, 50\%, and 60\%, and test the perplexity of the OPT and LLaMA2 models on the generation task datasets Alpaca, WikiText-2, PTB, and C4. For TRSP, we use the $\ell_2$-norm.}
\label{table12}
\adjustbox{width=0.6\textwidth}{
\begin{tabular}{cccccc}
\toprule
\textbf{Model} & \textbf{Pruning Ratio}  & \textbf{WikiText-2($\downarrow$)} & \textbf{Alpaca($\downarrow$)} & \textbf{PTB($\downarrow$)} & \textbf{C4($\downarrow$)} \\
\midrule
\multirow{7}{*}{\textbf{OPT-2.7B}} 
 & Dense      & 12.46  &11.64  & 17.97 & 14.32 \\
 & 10\%       & 12.78  &11.89  & 18.65 & 15.02 \\
 & 20\%       & 12.96  &12.15  & 19.38 & 16.36 \\
 & 30\%       & 15.52  &13.47  & 24.28 & 20.75 \\ 
 & 40\%       & 19.67  &15.82  & 32.25 & 25.41 \\
 & 50\%       & 25.62  &21.89  & 47.67 & 33.73 \\
 & 60\%       & 35.62  &32.19  & 59.87 & 47.38 \\
\midrule
\multirow{7}{*}{\textbf{OPT-6.7B}} 
 & Dense      & 10.85  &10.27  & 15.77 & 12.71 \\
 & 10\%       & 11.03  &10.95  & 16.78 & 13.46 \\
 & 20\%       & 11.48  &11.53  & 17.96 & 15.19 \\
 & 30\%       & 12.87  &12.75  & 20.86 & 17.52 \\ 
 & 40\%       & 14.87  &13.39  & 26.12 & 21.63 \\
 & 50\%       & 21.43  &16.51  & 35.36 & 28.74 \\
 & 60\%       & 29.63  &22.16  & 48.04 & 40.25 \\
\midrule
\multirow{7}{*}{\textbf{OPT-13B}} 
 & Dense      & 10.12  &9.46  &14.52 & 12.06 \\
 & 10\%       & 10.28  &9.89  & 15.12 & 12.73 \\
 & 20\%       & 10.39  &10.37 & 16.51 & 13.45 \\
 & 30\%       & 11.38  &11.12  & 19.42 & 16.14 \\ 
 & 40\%       & 12.67 & 12.33  & 24.62 & 20.79 \\
 & 50\%       & 15.38  & 14.76  & 30.88 & 27.65 \\
 & 60\%       & 28.23  & 20.98 & 39.38 & 36.14 \\
 \midrule
\multirow{7}{*}{\textbf{LLaMA2-7B}} 
 & Dense      & 5.47  &5.25 & 7.92 &7.26 \\
 & 10\%       & 5.58  &5.31 & 8.12 &7.47 \\
 & 20\%       & 6.13  &5.87 & 8.78 &7.92 \\
 & 30\%       & 8.26  &7.64 & 9.58 &8.85 \\ 
 & 40\%       & 10.28  & 9.79 & 12.87 & 11.42 \\
 & 50\%       & 14.58  &13.14  & 19.52 & 15.96 \\
 & 60\%       & 25.18 &21.46 & 30.14 & 28.32 \\
\midrule
\multirow{7}{*}{\textbf{LLaMA2-13B}} 
 & Dense      & 4.88  &4.63  & 7.16 & 6.73 \\
 & 10\%       & 4.99  &4.91  & 7.48 & 7.93 \\
 & 20\%       & 5.34  &5.26  & 8.25 & 8.75 \\
 & 30\%       & 6.87  &6.35  & 8.97 & 9.68 \\ 
 & 40\%       & 8.95  &7.92  & 10.08 & 11.26 \\
 & 50\%       & 11.23 &9.73  & 12.63 & 14.73\\
 & 60\%       & 15.74 &12.52 & 17.82 & 19.45 \\
\bottomrule
\end{tabular}}
\end{table*}

\subsection{The Accuracy of TRSP under Different Pruning Ratios on Zero-shot Tasks}
\label{E.2}
To systematically evaluate the performance of TRSP on zero-shot tasks across different pruning rates, we adopt the experimental setup outlined in Section \ref{4.2}, in which 128 samples are randomly sampled from the WikiText-2 training set to guide both the iterative learning of layer weights and the regularization process. We assess the accuracy of different model configurations at pruning rates of 10\%, 20\%, 30\%, 40\%, 50\%, and 60\% across a diverse set of benchmark datasets, including PIQA, WinoGrande, HellaSwag, ARC-e, and ARC-c. The results, summarized in Table \ref{table13}, provide insights into the impact of sparsity on zero-shot generalization. Notably, the analysis reveals that TRSP maintains competitive performance even at higher pruning rates, demonstrating its effectiveness in preserving reasoning and commonsense understanding across different tasks.
\begin{table*}[!htbp]
\centering
\caption{Accuracy comparison of TRSP with different pruning ratios. We set the pruning ratios to 10\%, 20\%, 30\%, 40\%, 50\%, and 60\%, and test the accuracy of the OPT and LLaMA2 models on the zero-shot task datasets PIQA, WinoGrande, HellaSwag, ARC-e and ARC-c. For TRSP, we use the $\ell_2$-norm. `Avg\_Acc' represents the average accuracy.}
\label{table13}
\adjustbox{width=\textwidth}{
\begin{tabular}{cccccccc}
\toprule
\textbf{Model} & \textbf{Pruning Ratio} & \textbf{PIQA(\%)} & \textbf{WinoGrande(\%)} & \textbf{HellaSwag(\%)} & \textbf{ARC-e(\%)}&\textbf{ARC-c(\%)}&\textbf{Avg\_Acc(\%)} \\
\midrule
\multirow{7}{*}{\textbf{OPT-2.7B}} 
 & Dense      & 74.81   &61.01 &60.58& 54.42&31.14 &56.39  \\
 & 10\%        &72.35  &60.75  &51.53 & 52.73&29.15 & 53.30 \\
 & 20\%        & 71.47  &60.46 &48.72 & 51.95&28.04& 52.13 \\
 & 30\%       & 66.73  &58.53  &44.36 & 50.02&26.58 &49.24  \\ 
 & 40\%       & 63.57  &55.32 &41.63 & 47.79& 25.24&46.71 \\
 & 50\%        & 58.48  &52.29  &40.52 & 46.26&21.32 & 43.77 \\
 & 60\%        & 52.85  &49.59  &38.46 &43.24&16.07 &40.04  \\
\midrule
\multirow{7}{*}{\textbf{OPT-6.7B}} 
 & Dense      & 76.39  &65.19  &67.16& 60.14& 34.64 & 60.70 \\
 & 10\%        & 75.42  &63.38  &65.16 & 57.23& 32.89& 58.82 \\
 & 20\%        & 74.58  &62.25  &61.46 & 55.98& 31.75&57.20  \\
 & 30\%       & 71.88  &60.56  &59.15 & 53.62&28.54 &54.75  \\ 
 & 40\%       & 66.98  &57.25  &53.66 & 49.42& 25.73&50.61 \\
 & 50\%        & 62.78  &54.56  &45.63 & 46.38&21.75 & 46.22 \\
 & 60\%        & 54.35  &50.26  &41.32 & 42.69&18.23 & 41.37 \\
\midrule
\multirow{6}{*}{\textbf{OPT-13B}} 
 & Dense      & 76.82  &64.80  &69.81 &61.87&35.67 & 61.79 \\
 & 10\%        & 75.49  &64.67  &69.26 & 61.73&35.54 & 61.34 \\
 & 20\%        & 74.89  &64.59  &68.95 & 61.62& 35.41& 61.09 \\
 & 30\%       & 71.46  &62.67  &66.53 & 59.39& 33.61&58.73  \\ 
 & 40\%       & 68.52  &60.39  &63.57 &56.12&28.83 & 55.49\\
 & 50\%        & 63.35  &57.62  &57.43 & 51.65&24.09 & 50.83 \\
 & 60\%        & 56.28  &53.06  &51.69 & 47.26& 21.87& 46.03 \\
 \midrule
\multirow{7}{*}{\textbf{LLaMA2-7B}} 
 & Dense      &79.11  &69.06 &75.99  & 74.58& 46.25&69.00  \\
 & 10\%        & 77.62 &68.45  &70.25 &69.85 &43.42 & 65.92 \\
 & 20\%        & 75.38  &67.96  &65.26 &65.83 &41.57 & 63.20 \\
 & 30\%       & 71.29  &64.77  &57.68 &60.45 &38.65 & 58.57 \\ 
 & 40\%       & 67.61  &60.38  &54.12 &57.53 &35.45 & 55.02\\
 & 50\%        & 61.87  &56.42  &51.62 & 53.45&30.28 & 50.73 \\
 & 60\%        & 55.39  &51.54  &48.73 &47.42 &27.08 &  46.03\\
\midrule
\multirow{7}{*}{\textbf{LLaMA2-13B}} 
 & Dense       & 80.47  &72.22  &79.39 & 77.48& 49.23 &71.76  \\
 & 10\%        & 78.45  &71.87  &74.32 & 76.26& 48.39& 69.86 \\
 & 20\%        & 76.12 &70.91   &70.06 & 74.68& 46.51& 67.66 \\
 & 30\%        & 72.54  &68.15  &63.51 & 70.88&43.21 & 63.66 \\ 
 & 40\%        & 71.45 &65.52   &62.28 & 69.13& 41.79& 62.03\\
 & 50\%        & 67.36 &58.52   &60.65 & 63.45&37.88 & 57.57 \\
 & 60\%        & 61.89 &55.26  &58.06 &  57.62&31.74 & 52.91 \\
\bottomrule
\end{tabular}}
\end{table*}
\section{Similarity Changes in Unregularized Layers}
\label{F}
We plot the input-output similarity of the unregularized layers in LLaMA2-7B, as described in Section~\ref{4.5}, in Figure~\ref{figure8}. According to the results shown below, the similarity of all layers except for the first one is already high before regularization is applied. After applying regularization, the similarity in these unregularized layers decreases, indicating that the transformations undergone by the inputs in these layers become more substantial. This suggests that more information is being captured in these layers compared to before, implying that the regularization process may cause information to shift from the regularized layers to the unregularized ones.
\begin{figure}[htbp]
	\centering
    \includegraphics[width=.99\columnwidth]{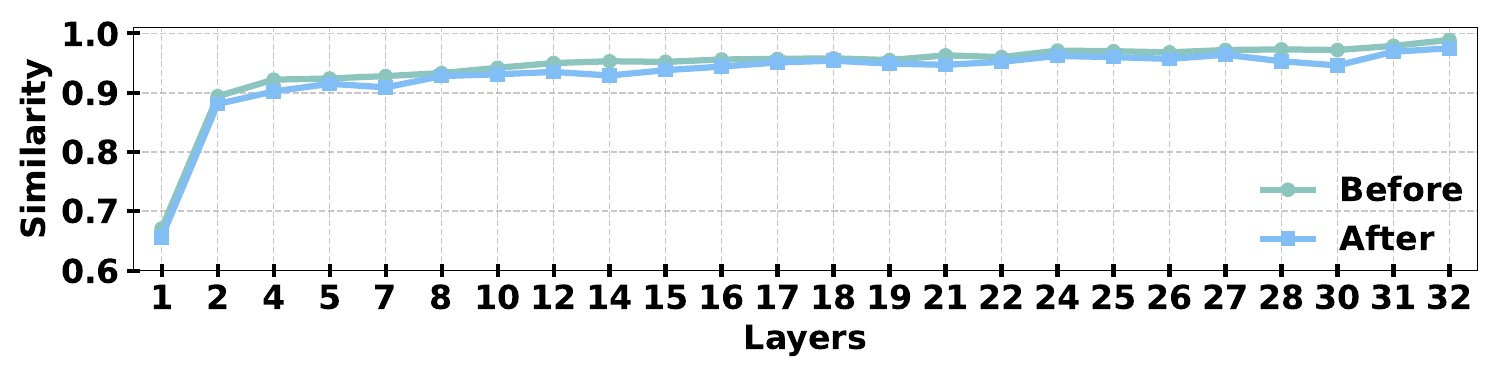}
    \caption{The input-output similarity of the unregularized layers.}
  \label{figure8}
\end{figure}

\section{Cosine Similarity}
\label{sec:appendix}
\begin{equation}
\begin{split}
\label{equation6}
\text{CosSim}(\mathbf{X_{\text{in}}^i},\mathbf{X_{\text{out}}^i})=\quad\quad\quad\quad\quad\quad\quad\quad\\\frac{1}{b n} \sum_{k=1}^{b} \sum_{j=1}^{n} \frac{ \left\langle \mathbf{X_{\text{in}}^i}[k,j,:],\ \mathbf{X_{\text{out}}^i}[k,j,:] \right\rangle }{ \left\| \mathbf{X_{\text{in}}^i}[k,j,:] \right\|_2 \cdot \left\| \mathbf{X_{\text{out}}^i}[k,j,:] \right\|_2 }
\end{split}
\end{equation}

\section{Experiments on Llama-2-70B}
\label{appendix:llama2-70b}

To further validate the scalability of our method, we conduct additional experiments on a larger-scale model, Llama-2-70B. All comparison methods are evaluated under the same experimental settings as described in the main paper. The results are summarized in Table~\ref{tab:llama2-70b}.

As shown in Table~\ref{tab:llama2-70b}, both TRSP-$\ell_1$ and TRSP-$\ell_2$ consistently outperform all baseline methods across the majority of evaluation benchmarks at a 25\% pruning ratio. Specifically, TRSP-$\ell_2$ achieves the lowest perplexity (4.13) and the highest average accuracy (74.32\%), while TRSP-$\ell_1$ attains the best performance on PIQA (81.45\%), ARC-e (77.97\%), and ARC-c (54.63\%). These results demonstrate that our method scales effectively to larger models and maintains a clear advantage over existing pruning approaches without requiring any retraining.

\begin{table*}[htbp]
\centering
\caption{Performance comparison of structured pruning methods on Llama-2-70B at a 25\% pruning ratio. \textbf{Bold} values indicate the best results among pruned models.}
\label{tab:llama2-70b}
\resizebox{\textwidth}{!}{%
\begin{tabular}{llccccccc}
\toprule
\textbf{Method} & \textbf{Pruning Ratio} & \textbf{PPL ($\downarrow$)} & \textbf{PIQA (\%)} & \textbf{WinoGrande (\%)} & \textbf{HellaSwag (\%)} & \textbf{ARC-e (\%)} & \textbf{ARC-c (\%)} & \textbf{Avg\_Acc (\%)} \\
\midrule
Dense           & 0\%  & 3.32 & 82.70 & 77.98 & 83.84 & 80.98 & 57.34 & 76.57 \\
\midrule
SLEB            & 25\% & 5.06 & 78.81 & 70.25 & 74.89 & 73.56 & 46.45 & 68.79 \\
ShortGPT        & 25\% & 4.85 & 80.15 & 75.62 & 79.85 & 76.41 & 50.29 & 72.46 \\
LaCo            & 25\% & 4.92 & 80.33 & 74.26 & 76.93 & 74.85 & 48.47 & 70.97 \\
Shortened LLaMA & 25\% & 4.98 & 79.52 & 73.15 & 76.37 & 75.15 & 47.89 & 70.42 \\
\midrule
TRSP-$\ell_2$   & 25\% & \textbf{4.13} & 81.35 & \textbf{76.73} & \textbf{81.26} & 77.69 & 54.56 & \textbf{74.32} \\
TRSP-$\ell_1$   & 25\% & 4.28 & \textbf{81.45} & 76.14 & 81.03 & \textbf{77.97} & \textbf{54.63} & 74.24 \\
\bottomrule
\end{tabular}%
}
\end{table*}

\section{Comparison with Different Granularity Pruning Methods}
\label{appendix:granularity-comparison}

To provide a more comprehensive evaluation, we further compare TRSP with state-of-the-art pruning methods of different granularities, including the unstructured pruning method Wanda~\cite{sun2024a} with a 2:4 sparsity pattern and the channel pruning method SliceGPT~\cite{ashkboos2024slicegpt}. For a fair comparison, the pruning ratio is set to 25\% for both TRSP and SliceGPT. All experiments are conducted on Llama-2-70B, and the data and settings are kept consistent with Section~4.2 of the main paper. Each method is evaluated using Perplexity (PPL) on Wikitext-2 and accuracy on several zero-shot benchmarks. The acceleration effect is measured by throughput improvement and inference speedup ratio. The results are reported in Table~\ref{tab:granularity-comparison}.

As shown in Table~\ref{tab:granularity-comparison}, TRSP achieves notably higher throughput improvement (1.38$\times$) and speedup ratio (1.36$\times$) compared to both Wanda and SliceGPT, while simultaneously maintaining superior task performance. Specifically, TRSP-$\ell_2$ attains the highest average accuracy (74.32\%) among all pruned models, surpassing Wanda by 1.55\% and SliceGPT by 3.89\%. Although Wanda with 2:4 sparsity yields a competitive average accuracy of 72.77\%, its throughput improvement (0.98$\times$) and speedup (1.10$\times$) are considerably lower, as unstructured sparsity patterns are less hardware-friendly. SliceGPT offers moderate acceleration (1.19$\times$ speedup) but suffers from a significant accuracy drop, particularly on HellaSwag (69.14\%). These results demonstrate that TRSP, as a layer pruning approach, achieves a more favorable trade-off between model performance and inference efficiency compared to state-of-the-art pruning techniques of other granularities.

\begin{table*}[htbp]
\centering
\caption{Comparison of pruning methods with different granularities on Llama-2-70B. \textbf{Bold} values indicate the best results among pruned models.}
\label{tab:granularity-comparison}
\resizebox{\textwidth}{!}{%
\begin{tabular}{llcccccccccc}
\toprule
\textbf{Method} & \textbf{Type} & \textbf{Pruning Ratio} & \textbf{PPL ($\downarrow$)} & \textbf{PIQA (\%)} & \textbf{WinoGrande (\%)} & \textbf{HellaSwag (\%)} & \textbf{ARC-e (\%)} & \textbf{ARC-c (\%)} & \textbf{Avg\_Acc (\%)} & \textbf{Throughput} & \textbf{Speedup} \\
\midrule
Dense           & /                      & 0\%  & 3.32 & 82.70 & 77.98 & 83.84 & 80.98 & 57.34 & 76.57 & /              & /          \\
\midrule
Wanda           & unstructured pruning   & 2:4  & 5.22 & 80.45 & 74.89 & 79.58 & 77.03 & 51.90 & 72.77 & 0.98$\times$   & 1.10$\times$ \\
SliceGPT        & channel pruning        & 25\% & 4.79 & 75.82 & 76.38 & 69.14 & 78.25 & 52.57 & 70.43 & 1.11$\times$   & 1.19$\times$ \\
\midrule
TRSP-$\ell_2$   & layer pruning          & 25\% & \textbf{4.13} & 81.35 & \textbf{76.73} & \textbf{81.26} & 77.69 & 54.56 & \textbf{74.32} & \textbf{1.38$\times$} & \textbf{1.36$\times$} \\
TRSP-$\ell_1$   & layer pruning          & 25\% & 4.28 & \textbf{81.45} & 76.14 & 81.03 & \textbf{77.97} & \textbf{54.63} & 74.24 & \textbf{1.38$\times$} & \textbf{1.36$\times$} \\
\bottomrule
\end{tabular}%
}
\end{table*}

\section{Theoretical Analysis of Knowledge Transfer in Second-Stage Regularization}
\label{appendix:knowledge-transfer}

In this section, we provide a theoretical analysis to explain why the second-stage regularization in TRSP facilitates knowledge transfer. The total loss function of TRSP is defined as:
\begin{equation}
    L_{\text{sum}} = L(W, X) + \lambda_2 \sum_{i=0}^{|P|-1} \left| X_{\text{out}}^{i} - X_{\text{in}}^{i} \right|,
    \label{eq:total-loss}
\end{equation}
which consists of two competing objectives:

\begin{enumerate}
    \item \textbf{Language modeling loss} $L(W, X)$: This term preserves the total amount of knowledge encoded in the model by maintaining its predictive capability on the training data.
    
    \item \textbf{Regularization term} $\lambda_2 \sum_{i=0}^{|P|-1} |X_{\text{out}}^{i} - X_{\text{in}}^{i}|$: This term encourages the input-output mapping $f_i$ of the layers targeted for pruning to approximate an \textbf{identity function}, i.e., $X_{\text{out}}^{i} \approx X_{\text{in}}^{i}$.
\end{enumerate}

These two terms impose conflicting constraints on the model. The regularization term drives the pruning-targeted layers toward identity mappings, effectively reducing the functional contribution of these layers. Meanwhile, the language modeling loss ensures that the overall model performance does not degrade during this process. To satisfy both objectives simultaneously, the model is compelled to redistribute the knowledge originally stored in the pruning-targeted layers to the remaining layers.

Empirically, we observe that prior to regularization, the knowledge resides in the discrepancy between the input and output of each layer, i.e., $X_{\text{out}}^{i} - X_{\text{in}}^{i} \neq 0$. After the second-stage regularization, this discrepancy is substantially reduced, indicating that the targeted layers have been driven toward identity transformations. Crucially, the language modeling loss remains nearly unchanged throughout this process, which confirms that the model's overall knowledge is preserved rather than lost. We therefore conclude that a knowledge transfer from the pruning-targeted layers to the remaining layers has occurred, enabling the subsequent removal of these layers with minimal performance degradation.

\end{document}